\documentclass[letterpaper, 10 pt, conference]{ieeeconf}  

\IEEEoverridecommandlockouts  
\usepackage{cite}

\usepackage{amsmath, amssymb, amsfonts, amsthm}
\usepackage{graphicx}
\usepackage{float}
\usepackage{textcomp}
\usepackage{subfig}
\usepackage{algorithm}
\usepackage{algpseudocode}
\usepackage[dvipsnames]{xcolor}
\usepackage{booktabs}
\usepackage{multirow}
\usepackage{siunitx}
\usepackage{mwe}

\usepackage[dvipsnames]{xcolor}

\usepackage{mathtools}
\DeclarePairedDelimiter\norm{\lVert}{\rVert}%

\usepackage[capitalize,noabbrev,nameinlink]{cleveref}
\crefformat{equation}{(#2#1#3)}
\Crefname{algocfline}{Algorithm}{Algorithms}
\Crefname{algocf}{line}{lines}
\Crefname{assumption}{Assumption}{Assumptions}
\crefrangeformat{assumption}{Assumptions~#3#1#4-#5#2#6}

\usepackage{amsmath, nccmath}
\usepackage{etoolbox}
\AtBeginEnvironment{bmatrix}{\setlength{\arraycolsep}{1pt}}
\usepackage{tabstackengine}
\setstackEOL{\cr}

\newcommand{\xdim}{n}
\newcommand{\tdim}{r}

\newcommand{\bxi}{\boldsymbol{\xi}}

\newcommand{\bXi}{\mathbf{\Xi}}
\newcommand{\bdX}{\mathbf{\dot{X}}}

\newcommand*{\vertbar}{\rule[-1ex]{0.5pt}{2.5ex}}

\DeclareMathOperator*{\argmin}{argmin}

\newcommand{\rotvert}{\rotatebox[origin=c]{90}{$\vert$}}
\newcommand{\rowsvdots}{\multicolumn{1}{@{}c@{}}{\vdots}}

\makeatletter
\let\oldnorm\norm
\def\norm{\@ifstar{\oldnorm}{\oldnorm*}}
\makeatother

\newtheorem{remark}{Remark}

\newtheorem*{proposition-non}{Proposition}
\newtheorem{problem}{Problem}

\graphicspath{ {./images/} } 

\title{\LARGE \bf
Symbolic Regression on Sparse and Noisy Data with Gaussian Processes  
} 

\author{
Junette Hsin$^{1}$, 
Shubhankar Agarwal$^{2}$, 
Adam Thorpe$^{1}$, 
Luis Sentis$^{1}$, and David Fridovich-Keil$^{1}$
\thanks{$^{1}$Departments of Aerospace Engineering \& Engineering Mechanics and $^{2}$Electrical and Computer Engineering,
        University of Texas at Austin, \{{\tt\small jhsin, somi.agarwal, adam.thorpe, lsentis, dfk\}@utexas.edu}}%
}
 
\begin{document}

\maketitle 


\thispagestyle{empty}
\pagestyle{empty}

\begin{abstract}

In this paper, we address the challenge of deriving dynamical models from sparse and noisy data. High-quality data is crucial for symbolic regression algorithms; limited and noisy data can present modeling challenges. To overcome this, we combine Gaussian process regression with 
a 
sparse identification of nonlinear dynamics (SINDy) method 
to denoise the data and identify nonlinear dynamical equations. 
%
%
Our approach GPSINDy offers improved robustness with sparse, noisy data compared to SINDy alone. 
We demonstrate its effectiveness on simulation data from Lotka-Volterra and unicycle models and hardware data from an NVIDIA JetRacer system.
We show superior performance over baselines including more than 50\% improvement 
over SINDy and other baselines  
in predicting future trajectories from noise-corrupted and sparse 5 Hz data.  


\end{abstract}


\section{Introduction}
\label{sec:intro}

Accurate dynamic models are crucial for effective robot design and operation. 
In many cases, it is desirable to obtain analytic expressions over black box models, as 
analytic models
extrapolate well beyond the training dataset and are more suitable for system analysis. 
One approach that has received significant attention is the Sparse Identification of Nonlinear Dynamics (SINDy) algorithm \cite{sindy_brunton}. 
It employs symbolic regression---a least-squares-based method---to learn 
the system dynamics purely from data 
using a predefined set of candidate functions. 
SINDy is simple in its approach but suffers from limitations in practice. 
%
%
In particular, 
the accuracy
of the learned solution relies heavily on the selection of proper candidate function terms, and measurement noise in the data can significantly degrade the performance of SINDy for even simple systems
\cite{cortiella_sindy_l1_regularized}. 
Additionally, SINDy requires derivative data, which is often obtained through finite differencing or other approximation methods which can add additional error \cite{wentz_derivative-based_2023}.
%
%
%
The sparsity of the data 
also impacts SINDy's performance; while it can identify models with limited data \cite{kaiser2018sparse}, low frequency data may reduce model accuracy. 
%
\textit{
In this work, we devise a method that combines Gaussian process (GP) regression in conjunction with SINDy 
to learn system dynamics using sparse and noisy data. 
%
}

SINDy's key advantage lies in discovering understand-
able models that balance accuracy and simplicity. 
While other data-driven methods have had success 
\cite{
rasmussen_gps 
}, 
limited data often limit their effectiveness, and the models they discover lack insight into the system's structure \cite{noauthor_ensemble-sindy_nodate}. 
In contrast, SINDy has been widely applied across various disciplines to understand the underlying structure of physical phenomena
\cite{
schaeffer_sindy_integral, 
boninsegna_sindy_stochastic, kaheman_sindy_pi}. 
%
In the field of robotics, it has been used to learn the dynamics of actuated systems for control purposes 
such as modeling jet engines for feedback linearization and sliding mode control \cite{l2020modeling}. 
SINDy's appeal lies in its simple and highly adaptable sparse linear regression approach, requiring less data compared to methods like neural networks
\cite{noauthor_ensemble-sindy_nodate}. 

However, 
noise in the data remains a problem. 
A growing body of research based on SINDy seeks to mitigate the impact of noise on identifying the correct system dynamics. 
Reactive SINDy \cite{hoffmann2019reactive} uses vector-valued functions with SINDy to uncover underlying biological cell structures 
from noisy data, but 
can fail to converge to the correct reaction network with increasing levels of noise. 
%
PiDL-SINDy \cite{chen2021physics} utilizes a physics-informed neural network with sparse regression to learn the system dynamics, and DSINDy \cite{wentz_derivative-based_2023} and a modified SINDy using automatic differentiation (AD) \cite{kaheman_automatic_2020} simultaneously de-noise the data and identify the governing equations. 
However, PiDL-SINDy \cite{chen2021physics} and AD-SINDy \cite{kaheman_automatic_2020} run into computational bottlenecks and challenges with the structure of their optimization problems. 
Derivative-based approaches show promise, but DSINDy makes assumptions on the structure of its function library that may not be true in practice.  
ESINDy \cite{noauthor_ensemble-sindy_nodate} proposes 
a statistical framework to compute the 
probabilities of candidate functions
from an ensemble of models identified from noisy data, but its approach relies on 
Sequentially Thresholded Least Squares (STLS) regression. 
%
STLS can be effective for small levels of noise, but it deteriorates with larger noise levels
\cite{cortiella_sindy_l1_regularized}. 
%

Advancements in non-parametric based approaches 
also tackle the problem of learning governing equations from noisy data. 
Neural networks have been used to parameterize the state derivatives through a blackbox differential equation solver \cite{chen2018neural}, 
and Gaussian processes have been used to infer parameters of linear equations from scarce and noisy observations \cite{Raissi_ML_GP} and generate vector fields of nonlinear differential equations \cite{heinonen2018learning}. 
Gaussian process regression is particularly effective as an interpolation tool and at reducing the noise in measurement data \cite{rasmussen_gps}.

%
%
%
\textbf{Contributions: }
First, the novelty of our method lies in how we approximate the state derivatives by constructing the Gaussian process kernel using smoothed state measurements. 
%
%
This approach addresses issues caused by measurement noise and low temporal resolution for model learning. 
%
We learn the relationship between the state and time derivatives using Gaussian processes, and then determine analytic expressions for the dynamics with SINDy. 
%
%
Second, we benchmark our method on both simulated and experimental hardware data, comparing it with SINDy, neural network models, and other baselines. The results show that our approach is notably more robust in identifying models from sparse and noisy data.

\section{Problem Formulation}
\label{sec:prob_form}

Consider a system characterized by unknown dynamics 
\begin{align}
    \label{eq:prob}
    \dot{\boldsymbol{x}}(t) &= f( \boldsymbol{x}(t), \boldsymbol{u}(t) ), 
\end{align} 
where $t \in \mathbb{R}$, $\boldsymbol{x}(t) \in \mathbb{R}^n$ denotes the state of the system at time $t$, 
and $\boldsymbol{u}(t) \in \mathbb{R}^m$ the control input. 
%
We presume that $f : \mathbb{R}^{n} \times \mathbb{R}^{m} \to \mathbb{R}^{n}$ in \eqref{eq:prob} is unknown, but composed of a sparse set of terms that contribute to the system dynamics, e.g. damping or inertial forces \cite{wentz_derivative-based_2023}.
%
%

We assume that we have access to a dataset $\boldsymbol{X}$ consisting of a sequence of $r \in \mathbb{N}$ state measurements corrupted by noise and control inputs $\boldsymbol{U}$ 
taken at discrete times 
$t_{1}, t_{2}, \ldots, t_{r}$,
given by 
\begin{equation}
\label{eq:dataset}
    \begin{split}
        \boldsymbol{X} & = \lbrace \boldsymbol{x}(t_1) + \boldsymbol{\epsilon}_{1}, \boldsymbol{x}(t_2) + \boldsymbol{\epsilon}_{2}, \ldots, \boldsymbol{x}(t_r) + \boldsymbol{\epsilon}_{r} \rbrace \\ 
        \boldsymbol{U} & =  \lbrace \boldsymbol{u}(t_{1}), \boldsymbol{u}(t_2), \ldots, \boldsymbol{u}(t_r) \rbrace , 
    \end{split}
\end{equation} 
where $\boldsymbol{\epsilon}_{i} \sim \mathcal{N}(0, \delta^2 I)$.
The derivatives of the state with respect to time 
$\dot{\boldsymbol{x}}(t) \in \mathbb{R}^\xdim$ 
are not directly measurable and must be approximated from $\boldsymbol{X}$,
e.g. using (central) finite differencing.
Let $ \dot{\boldsymbol{X}}$ be the approximate state derivatives with respect to time of the points in the dataset $\boldsymbol{X}$ in \eqref{eq:dataset}. 
%
Intuitively, we can view $\boldsymbol{X}$ and $\dot{\boldsymbol{X}}$ as matrices in $\mathbb{R}^{ \tdim \times n }$
where the $i^{\rm th}$ row $ \boldsymbol{X}_i $ corresponds to the state at time $t_i$ 
\begin{equation}
    \boldsymbol{X} = 
    \begin{bmatrix} 
        & \rotvert & \boldsymbol{X}_1 & \rotvert &    \\ 
        &          & \rowsvdots       &          &    \\ 
        & \rotvert & \boldsymbol{X}_r & \rotvert &     
    \end{bmatrix},
    \label{eq:X_data}
\end{equation}
and the $i^{\rm th}$ row $\dot{\boldsymbol{X}}_{i}$ is the time derivative of $\boldsymbol{X}_{i}$.
%
The state measurements $\boldsymbol{X}$, affected by noise, 
lead to inaccurate estimates of the time derivatives in $ \dot{ \boldsymbol{X} } $. 

We assume that the dynamics can be described by a linear combination of relatively few basis function terms such as polynomials of varying degrees, sinusoidal terms, or exponential functions. 
For instance, 
\begin{equation}
    \dot{\boldsymbol{x}}(t) = \boldsymbol{\Theta}(\boldsymbol{x}(t), \boldsymbol{u}(t))^{\top} \boldsymbol{\Xi},
\end{equation}
where $\boldsymbol{\Theta}(\boldsymbol{x}(t), \boldsymbol{u}(t)) \in \mathbb{R}^{p}$ is the candidate function library of basis functions evaluated at the current state $\boldsymbol{x}(t)$ and control input $\boldsymbol{u}(t)$ and $\boldsymbol{\Xi} \in \mathbb{R}^{p \times n}$ is a matrix of real-valued coefficients that weigh the candidate function terms. 
For simplicity, using the dataset $\boldsymbol{X}$, the applied control inputs $\boldsymbol{U}$, and the state derivatives $\dot{\boldsymbol{X}}$, we can write the relationship between the datasets via
\begin{equation}
    \dot{\boldsymbol{X}} = \boldsymbol{\Theta}(\boldsymbol{X}, \boldsymbol{U})^{\top} \boldsymbol{\Xi}.
    \label{eq:Xdot_theta_Xi}
\end{equation}
In practice, \eqref{eq:Xdot_theta_Xi} does not exactly hold as 
the dataset $\boldsymbol{X}$ is corrupted by noise, and the approximation of 
$ \dot{\boldsymbol{X}} $ 
introduces additional error as given by 
\begin{equation}
    \dot{\boldsymbol{X}} = \boldsymbol{\Theta(X},\boldsymbol{U})^{\top} \boldsymbol{\Xi} + \eta \boldsymbol{Z} ,   
    \label{eq:deriv_meas}
\end{equation} 
where $ \boldsymbol{Z} $ is a matrix of independent, identically distributed zero-mean Gaussian entries and $ \eta $ is the magnitude of the standard deviation of the noise. 

To find $ \boldsymbol{\Xi} $, one can use ordinary least-squares with 
$ \boldsymbol{X} $ and $ \dot{\boldsymbol{X}} $ 
to find the model $f$ from \eqref{eq:prob}. 
However, this approach does not lead to a sparse representation, instead overfitting the model to the data and finding a solution with nonzero elements in every element of $ \boldsymbol{\Xi} $. 
Sparsity is desirable to prevent overfitting, particularly with noisy data. 
%
Fortunately, 
LASSO \cite{tibshirani_lasso} has been shown to work well with noise, using $L_{1}$ regularization to promote sparsity in the solution. 
%


\begin{problem}[LASSO for Symbolic Regression] 
 We seek to solve the LASSO problem 
\begin{equation}
    \bxi_j = \argmin_{\boldsymbol{\xi} \in \mathbb{R}^{p}} \norm*{ \boldsymbol{\Theta}(\boldsymbol{X}, \boldsymbol{U})^{\top} \boldsymbol{\xi} - \dot{\boldsymbol{X}}_j  }_2 + \lambda \norm*{ \bxi }_1 , 
    \label{eq:sindy} 
\end{equation}
where the optimization variable $ \bxi_j \in \mathbb{R}^{p} $ is the $j^{\rm th}$ column of $ \bXi $ from \eqref{eq:deriv_meas}, $ \dot{\boldsymbol{X}}_j $ is the $j^{\rm th}$ column of $ \dot{\boldsymbol{X}} $ from \eqref{eq:deriv_meas}, and $ \lambda > 0$ is the $L_{1}$ regularization parameter.
\end{problem}
However, the $L_1$ regularization parameter $\lambda$ must be carefully chosen, which balances model complexity (determined by the number of nonzero coefficients in $\bXi$) with accuracy. 
%
%
Additionally, low frequency data 
might not provide enough information to learn models with rapidly evolving dynamics. 
Addressing the challenges posed by noise and data sparsity is crucial for accurate system identification.


\section{Approach}
\label{sec:methods}

Several techniques exist for de-noising data including Fourier transforms, a range of filtering methods, 
and neural networks 
\cite{vaseghi2008advanced}. 
In this work, we propose using Gaussian process (GP) regression to mitigate noise-related issues and enhance the analytical model’s precision. 

\textbf{Smoothing $\boldsymbol{X}$:} Like SINDy, GP regression yields a model for relating input and output data. 
Unlike SINDy, GP regression is \emph{non}-parametric; 
it models a probability distribution of the data $\boldsymbol{X}$ as a function of $\boldsymbol{t}$. 
This distribution is described by a mean function $m( \cdot )$ and covariance kernel function $k( \cdot,\cdot )$, and 
the negative log-likelihood 
of $\boldsymbol{X}$ is given by
\begin{equation}
    \begin{aligned}
    -\operatorname{log} p(\boldsymbol{X})  = & \, 
    \frac{1}{2}  
    \boldsymbol{X}^T (K + \theta_n^2)^{-1} \boldsymbol{X} \\ 
    & 
    + \frac{1}{2} \operatorname{log} | K + \theta_n^2 | 
    + \frac{n}{2} \operatorname{log} (2 \pi),  
    \end{aligned}
    \label{eq:gp_marg_likelihood}
\end{equation} 
where $K = k( \boldsymbol{t}, \boldsymbol{t} )$ and $\theta_n$ is the tunable noise variance hyperparameter.
GP regression performance is sensitive to the kernel choice; different kernels can lead to poor extrapolation, where the predicted mean reverts to the training dataset's mean function \cite{rasmussen_gps}. 
For example, the standard squared-exponential (SE) kernel is given by 
\begin{equation}
    k(  t_i, t_j ) = \theta_f^2 \exp \big( - \frac{1}{2 \theta_l ^2} || t_i - t_j || ^2 \big), 
\end{equation}
where $\theta_f$ is the signal variance hyperparameter and $\theta_l$ is the length scale hyperparameter. 
The SE hyperparameters are determined by minimizing \eqref{eq:gp_marg_likelihood} with respect to $\theta_f$, $\theta_l$, and $\theta_n$. 
Numerous kernel functions exist for fitting data with different structures, including the periodic, Mat\'ern, and Rational Quadratic (RQ) kernels. 
%
%
To determine which kernel best matches the data, we can evaluate its marginal log-likelihood;  
the lower the negative marginal log-likelihood, the more likely the kernel fits the data. 

To use GPs as a de-noising tool, 
we first assume that $ \boldsymbol{X} $ was generated at training times $\boldsymbol{t}$ from a zero-mean GP; it is common practice to assume a mean function of zero \cite{rasmussen_gps}.
Now, let $ \boldsymbol{X}_* $ be a random Gaussian vector generated from a GP at desired test times $ \boldsymbol{t}_* $ 
\begin{equation}
    \begin{bmatrix}
        \boldsymbol{X}_{*} \\
        \boldsymbol{X} 
        \end{bmatrix}\sim \mathcal{N} \left(\begin{bmatrix}
        0 \\
        0
        \end{bmatrix},\begin{bmatrix}
        K(\boldsymbol{t}_*, \boldsymbol{t}_*) & K(\boldsymbol{t}_*, \boldsymbol{t}) \\
        K(\boldsymbol{t}, \boldsymbol{t}_*)   & K(\boldsymbol{t},\boldsymbol{t}) + \theta_n^2 I 
    \end{bmatrix}\right), 
\end{equation}
where $r$ is the number of training points, and $r_*$ is the number of test points. 
$K(\boldsymbol{t},\boldsymbol{t}_*)$ denotes the $ r \times r_* $ matrix of the covariances evaluated at all pairs of training and test points, $K(\boldsymbol{t},\boldsymbol{t})$ is a $r \times r$ matrix of covariances, and likewise for $ K(\boldsymbol{t}_*, \boldsymbol{t}) $ and $K(\boldsymbol{t}_*, \boldsymbol{t}_*)$.  
To obtain smoothed estimates of $ \boldsymbol{X} $ evaluated at the test points, we condition the distribution of the training data on the test data to compute the posterior
\begin{equation}
    \boldsymbol{X}_{GP} = K(\boldsymbol{t}_*, \boldsymbol{t}) [ K(\boldsymbol{t},\boldsymbol{t}) + \theta_n^2 I ]^{-1} \, \boldsymbol{X} .
    \label{eq:X_GP}
\end{equation}

\textbf{Smoothing $\dot{\boldsymbol{X}}$:} Similarly to $\boldsymbol{X}$, the kernel for the state derivatives $ \dot{\boldsymbol{{X}}} $ can also be determined from its negative marginal log-likelihood 
\begin{equation}
    \begin{aligned}
    -\operatorname{log} p(\dot{\boldsymbol{X}})  = & \, 
    \frac{1}{2}  
    \dot{\boldsymbol{X}}^{\top} (K + \sigma_n^2 )^{-1} \dot{\boldsymbol{X}} 
    \\ 
    & 
    + \frac{1}{2} \operatorname{log} | K | 
    + \frac{n}{2} \operatorname{log} (2 \pi). 
    \end{aligned}
    \label{eq:Xdot_marg_likelihood}
\end{equation} 
by choosing the kernel function that minimizes \eqref{eq:Xdot_marg_likelihood}. 
Constructing a kernel which represents the structure of the data being modelled is a crucial aspect of using GPs. 
For the data $\dot{\boldsymbol{X}}$, we assume that the kernel $K = k( \cdot, \cdot )$ is not a function of $\boldsymbol{t}$, but a function of the state $\boldsymbol{X}$ or control input $\boldsymbol{U}$. 
For instance, the joint distribution of 
the training and test points for 
the state derivatives $ \dot{ \boldsymbol{X} } $ can be given by 
\begin{equation}
\begin{bmatrix}
\dot{\boldsymbol{X}}_{*} \\
\dot{\boldsymbol{X}}
\end{bmatrix} 
\! \sim  
\mathcal{N} 
\Bigg( 
\! 
\begin{bmatrix}
0 \\ 
0 
\end{bmatrix}\! , \! 
\begingroup 
\setlength\arraycolsep{1pt}
\begin{bmatrix}
K(\boldsymbol{X}_{*}, \boldsymbol{X}_{*}) \! & K(\boldsymbol{X}_{*}, \boldsymbol{X}) \\
K(\boldsymbol{X}, \boldsymbol{X}_{*})  \! &  K(\boldsymbol{X},\boldsymbol{X}) + \sigma_n^2 I 
\end{bmatrix}
\! 
\bigg) ,
\endgroup
\end{equation}
where $K(\boldsymbol{X},\boldsymbol{X}_*)$ denotes the $ r \times r_* $ matrix of covariances and similarly for $K(\boldsymbol{X},\boldsymbol{X})$, $ K(\boldsymbol{X}_*, \boldsymbol{X}) $ and $K(\boldsymbol{X}_*, \boldsymbol{X}_*)$. 
$ \sigma_n $ is the noise variance hyperparameter for $ \dot{\boldsymbol{X}} $. 
%
%
%
To calculate smoothed estimates of $\dot{\boldsymbol{X}}$ evaluated at the test points $\boldsymbol{X}_*$, we compute the posterior mean via 
\begin{equation}
    \dot{\boldsymbol{X}}_{GP} = K(\boldsymbol{X}_{*}, \boldsymbol{X}) [ K(\boldsymbol{X},\boldsymbol{X}) + \sigma_n^2 I ]^{-1} \, \dot{\boldsymbol{X}} .
    \label{eq:Xdot_GP}
\end{equation}
We note that $ \dot{\boldsymbol{X}}_{GP} $ can also be computed by using the smoothed state measurements $ \boldsymbol{X}_{GP} $ 
\begin{equation}
    \dot{\boldsymbol{X}}_{GP} = K(\boldsymbol{X}_{GP*}, \boldsymbol{X}_{GP}) [ K(\boldsymbol{X}_{GP},\boldsymbol{X}_{GP}) + \sigma_n^2 I ]^{-1} \, \dot{\boldsymbol{X}} , 
    \label{eq:Xdot_GP_X_GP}
\end{equation}
or by using the control inputs $\boldsymbol{U}$ as the kernel input 
\begin{equation}
    \dot{\boldsymbol{X}}_{GP} = K(\boldsymbol{U}, \boldsymbol{U}) [ K(\boldsymbol{U},\boldsymbol{U}) + \sigma_n^2 I ]^{-1} \, \dot{\boldsymbol{X}}.    
    \label{eq:Xdot_GP_U}
\end{equation}
Now that we have shown how to obtain  
$ \boldsymbol{X}_{GP} $ and $ \dot{\boldsymbol{X}}_{GP} $, we move forward to solve the problem in \eqref{eq:sindy}. 
\begin{remark}
The novelty of our method lies in how 
we use $ \boldsymbol{X}_{GP} $ or $ \boldsymbol{U} $ as the kernel input for estimating $ \dot{\boldsymbol{X}} $ to capture the structure present in the true dynamics. 
This approach can produce smoothed and more accurate state derivatives $ \dot{\boldsymbol{X}}_{GP} $ for symbolic regression. 
\end{remark} 

\textbf{LASSO and cross-validation:} We update the LASSO problem from \eqref{eq:sindy} to use 
$ \boldsymbol{X}_{GP} $ 
and 
$ \dot{\boldsymbol{X}}_{GP} $ when solving for the coefficients of the system dynamics 
\begin{equation}
    \bxi_j = \argmin_{\boldsymbol{\xi} \in \mathbb{R}^{p}} \norm*{ \boldsymbol{\Theta}(\boldsymbol{X}_{GP}, \boldsymbol{U})^{\top} \boldsymbol{\xi} - \dot{\boldsymbol{X}}_{GP, j}  }_2 + \lambda \norm*{ \bxi }_1 ,  
    \label{eq:gpsindy} 
\end{equation} 
where $ \dot{\boldsymbol{X}}_{GP, j} $ represents the $j^{\rm th}$ column of $ \dot{\boldsymbol{X}}_{GP} $. 

LASSO can be computationally expensive for large data sets. Fortunately, the objective function in \eqref{eq:gpsindy} is separable, making it suitable for 
optimization 
via splitting methods. One such method, the Alternating Direction Method of Multipliers (ADMM) \cite{boyd_admm}, 
is an algorithm that  
%
solves convex optimization problems splitting its primary variables into two parts, each of which is then updated in an alternating fashion. 
We can solve the LASSO problem in \eqref{eq:gpsindy} using ADMM by treating $ \boldsymbol{\bxi}_j $ as the primary variable to be split, resulting in 
\begin{equation}
    \begin{aligned}
    \bxi_j = \argmin_{\bxi, \boldsymbol{z} \in \mathbb{R}^{p}} & \; \; \norm*{ \boldsymbol{\Theta}(\boldsymbol{X}_{GP}, \boldsymbol{U})^{\top} \boldsymbol{\xi} - \dot{\boldsymbol{X}}_{GP, j}  }_2 + \lambda \norm*{ \boldsymbol{z} }_1 , \\ 
    \text{s.t.} & \; \; \bxi - \boldsymbol{z} = \boldsymbol{0}, 
    \end{aligned}
    \label{eq:admm}
\end{equation} 
where $\boldsymbol{z}$ is split from the primary variable. 
Full implementation details for ADMM can be found in \cite{boyd_admm}. 
%
%

%
Finally, we find a suitable $ \lambda $ that balances model complexity and accuracy. 
Cross-validation determines the optimal sparsity parameter $\lambda$ and hyperparameters by evaluating model performance across various training and test sets to achieve the best results  \cite{ito2003optimizing}. 
In this work, we perform cross-validation with LASSO to achieve a sparse solution with the best model fit based on the dataset. 

\subsection{GPSINDy: Symbolic Regression with GP Denoising} 
%
We evaluate the marginal log-likelihood of the state $\boldsymbol{X}$ using the SE, Mat\'ern 1/2, Mat\'ern 3/2, periodic, and RQ kernels via Equation \eqref{eq:gp_marg_likelihood} and optimize the hyperparameters for each respective kernel. 
We select the kernel with the lowest negative marginal log-likelihood score to compute $\boldsymbol{X}_{GP}$ via Equation \eqref{eq:X_GP}.  

For the state derivatives $\dot{\boldsymbol{X}}$, we construct the marginal likelihood using Equation \eqref{eq:Xdot_marg_likelihood} with the same aforementioned kernel functions as $\boldsymbol{X}$, but using  $\dot{\boldsymbol{X}}_{GP}$ and $\boldsymbol{U}$ as the kernel inputs. 
We choose the kernel with the lowest negative log-likelihood score and then compute the posterior mean $ \dot{\boldsymbol{X}}_{GP} $ using Equation \eqref{eq:Xdot_GP_X_GP} or \eqref{eq:Xdot_GP_U}. 
%
%

%

Finally, we use ADMM to solve the LASSO problem in \eqref{eq:gpsindy} to discover the dynamics for an unknown system using noisy measurements, and we call this method GPSINDy. 
\begin{algorithm} 
\begin{small}
\caption{ 
GPSINDy with ADMM (LASSO) 
}\label{alg:GPSINDy}
\begin{algorithmic}[1]
    \Require state measurements $\boldsymbol{X}$, control inputs $ \boldsymbol{U} $, computed state derivatives $\bdX$, $L_1$ parameter $ \lambda $
    \Ensure coefficients for active nonlinear terms $ \bXi $
    \State compute $\boldsymbol{X}_{GP}$ using \eqref{eq:X_GP} and $\dot{\boldsymbol{X}}_{GP}$ using  \eqref{eq:Xdot_GP_X_GP} or \eqref{eq:Xdot_GP_U}
    \State construct data matrix $ \boldsymbol{\Theta} $ using $ \boldsymbol{X}_{GP} $ and $ \boldsymbol{U} $  
    \For{ k = 1, 2, \dots, \xdim } 
        \State $\bxi_k = \operatorname{ADMM} ( \boldsymbol{\Theta}, \dot{ \boldsymbol{X} }_k, \lambda ) $
    \EndFor 
    \State $ \bXi = [ \bxi_1, \bxi_2, \dots , \bxi_{\xdim} ] $
\end{algorithmic} 
\end{small}
\end{algorithm} 



\section{Experiments \& Results}
\label{sec:results}

\begin{table}[!t]
\centering
\begin{tabular}{lllllll}
    \toprule
     & \multicolumn{2}{c}{Ground Truth} & \multicolumn{2}{c}{SINDy} & \multicolumn{2}{c}{GPSINDy (Ours)} \\
    \cmidrule(r){2-3} \cmidrule(r){4-5} \cmidrule(r){6-7}
    $\Theta$ term & $\dot{x}_{1}$ & $\dot{x}_{2}$ & $\dot{x}_{1}$ & $\dot{x}_{2}$ & $\dot{x}_{1}$ & $\dot{x}_{2}$ \\
    \midrule
    $x_{1}$                 & 1.1  & 0.0  & 1.108           & \textbf{0.0}      & \textbf{1.097}    & \textbf{0.0} \\
    $x_{2}$                 & 0.0  & -0.1 & \textbf{0.0}    & \textbf{-0.997}   & \textbf{0.0}      & -0.980 \\
    $x_{1}x_{2}$            & -0.4 & 0.4  & \textbf{-0.397} & 0.382             & -0.358            & \textbf{0.396} \\
    $x_{2}x_{2}$            & 0.0  & 0.0  & \textbf{0.0}    & \textbf{0.0}      & \textbf{0.0}      & -0.005 \\
    $\cos(x_{1})$           & 0.0  & 0.0  & \textbf{0.0}    & 0.016             & -0.049            & \textbf{0.0} \\
    $x_{1}\cos(x_{1})$      & 0.0  & 0.0  & \textbf{0.0}    & -0.005            & \textbf{0.0}      & \textbf{0.0} \\
    $x_{1}x_{1}\sin(x_{2})$ & 0.0  & 0.0  & -0.003          & \textbf{0.0}      & \textbf{0.0}      & \textbf{0.0} \\
    $x_{1}x_{2}\sin(x_{1})$ & 0.0  & 0.0  & \textbf{0.0}    & -0.007            & \textbf{0.0}      & \textbf{0.0} \\
    $x_{1}x_{2}\sin(x_{2})$ & 0.0  & 0.0  & 0.003           & \textbf{0.0}      & \textbf{0.0}      & \textbf{0.0} \\
    $x_{1}x_{1}\cos(x_{1})$ & 0.0  & 0.0  & \textbf{0.0}    & -0.009            & \textbf{0.0}      & \textbf{-0.003} \\
    $x_{1}x_{1}\cos(x_{2})$ & 0.0  & 0.0  & -0.001          & \textbf{0.0}      & \textbf{0.0}      & \textbf{0.0} \\
    \bottomrule
\end{tabular} 
\caption{
\small{\textbf{GPSINDy learns better coefficients over SINDy for the predator-prey model.} 
The bold values show the best learned coefficients compared to the ground-truth for both $\dot{x}_1$ and $\dot{x}_2$. }}
\label{tab:gpsindy_coeffs_compare}
\vspace{-2em}
\end{table}

We demonstrate the effectiveness of GPSINDy on the Lotka-Volterra and unicycle models and data from the NVIDIA JetRacer, 
benchmarking GPSINDy against SINDy \cite{sindy_brunton} and a neural network-based method, NNSINDy. 
NNSINDy uses a neural network (NN) for data refinement with 32 hidden neurons, trained using the ADAM optimizer, and LASSO for symbolic regression \cite{Kingma2015AdamAM}.
We refer to the ground-truth coefficients as $\boldsymbol{\Xi}_{\text{GT}}$ and the learned coefficients as $\boldsymbol{\Xi}_{\text{Learned}}$ for all experiments.

\textbf{Experimental setup: } 
Unless otherwise specified, 
we simulate all systems for $30 \si{\second}$ at  
discrete time steps $ t \in \{ t_1, t_2, \dots, t_\tdim \} $ with a $0.1 \si{\second}$ sampling interval. 
We compute the derivatives $ \dot{\boldsymbol{x}}(t) $ using the true dynamics and add
$\boldsymbol{\epsilon} \sim \mathcal{N}\left(0, \sigma^2 \right)$ 
on top of $\boldsymbol{x}(t)$ and $ \dot{\boldsymbol{x}}(t) $ to simulate measurement noise, thereby obtaining $ \boldsymbol{X} $ and $ \dot{\boldsymbol{X}} $. 
We use the first 80\% of the simulated data for training and the rest for validation. 
We smooth $ \boldsymbol{X} $ using \eqref{eq:X_GP} and $ \dot{\boldsymbol{X}} $ using \eqref{eq:Xdot_GP} or \eqref{eq:Xdot_GP_X_GP} to obtain $ \boldsymbol{X}_{GP} $ and $ \dot{\boldsymbol{X}}_{GP} $ from the training data. 

\begin{figure}[!t]
    \centering
    \includegraphics[width=0.45\textwidth]{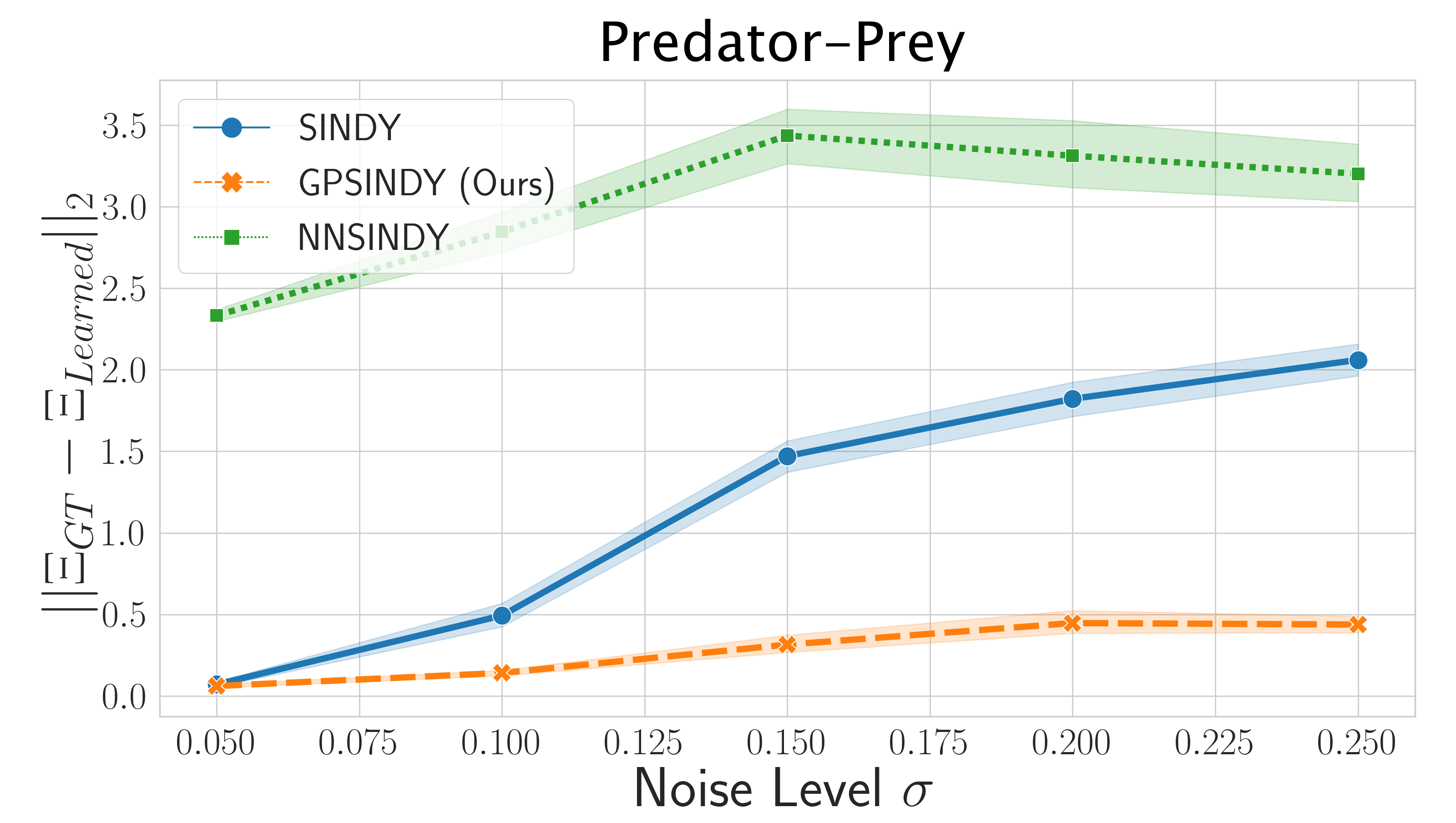}
    \includegraphics[width=0.45\textwidth]{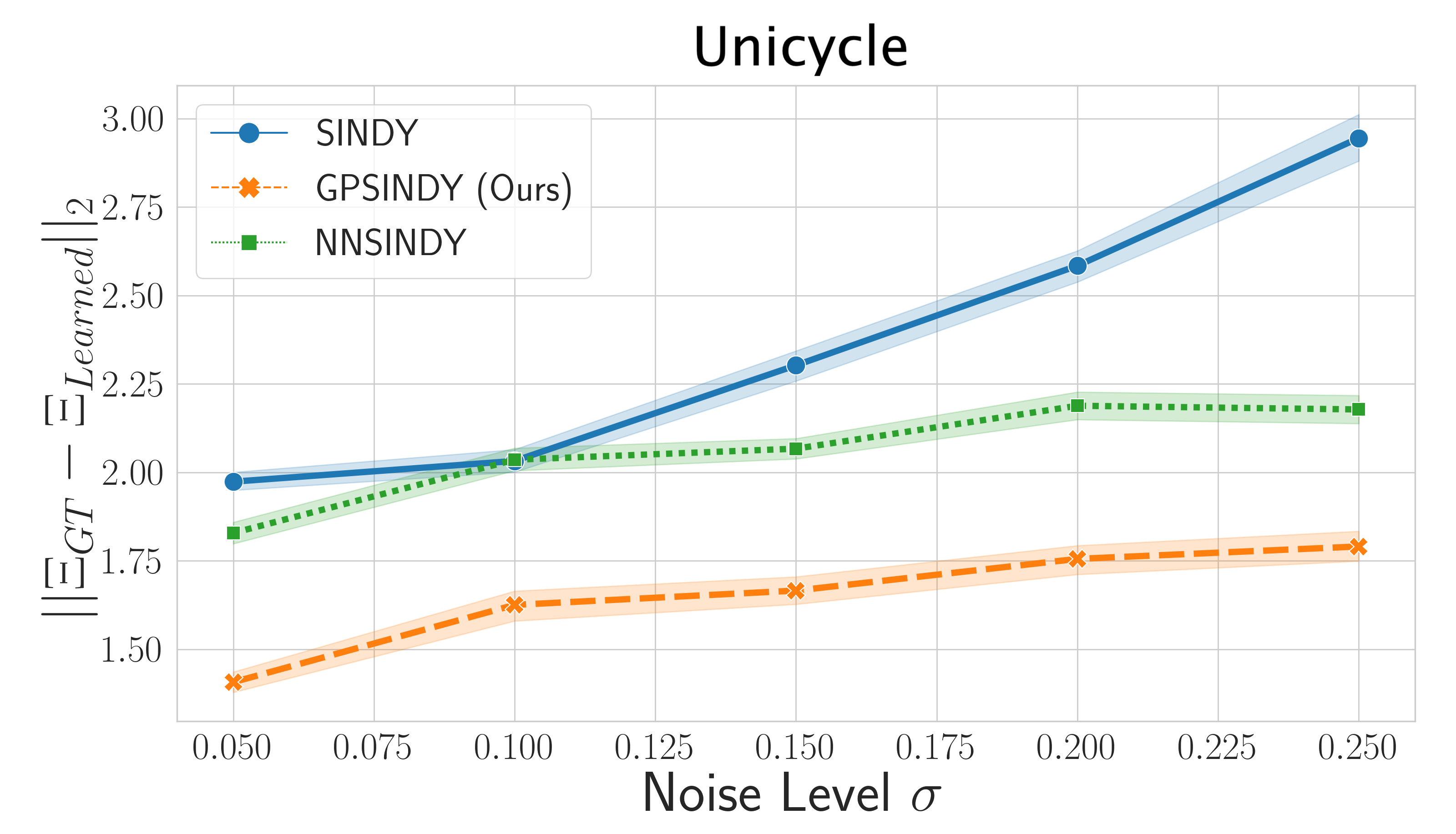}
    \vspace{-0.5em}
    \caption{
    \small{\textbf{GPSINDy achieves lowest error learning model coefficients for the predator-prey and unicycle models}. Contrast the dynamics learned by SINDy (blue), GPSINDy (orange), and NNSINDy (green) with trials repeated over 40 seeds for each noise standard deviation $ \sigma $, which varies from $0.050$ to $0.25$. The vertical axis represents the mean-squared error between the ground-truth ($\boldsymbol{\Xi}_{\text{GT}}$) and the learned coefficients ($\boldsymbol{\Xi}_{\text{Learned}}$). The ribbon indicates the standard deviation around the mean line; lower is better.
    }}
\label{fig:predator_prey_unicycle_err_coeff}
\vspace{-2em}
\end{figure}

We compute $ \boldsymbol{\Theta}( \boldsymbol{X}_{GP} ) $ using \eqref{eq:exp_Thetax} and 
solve the 
problem in \eqref{eq:admm} using $\lambda = 0.1$. 
For the baseline NNSINDy, we train a NN to predict $\dot{\boldsymbol{X}}$ given the observations of $\boldsymbol{X}$ using the training data and apply LASSO regression to discover the coefficients. 
We choose the candidate function library $\boldsymbol{\Theta}( \boldsymbol{X}, \boldsymbol{U} ) $ (unless otherwise specified) such that it consists of polynomial terms up to $3^{\rm rd}$ order, $\sin$ and $\cos$ terms, and products of the 
aforementioned 
terms. 
For example, 
%
\begin{equation}
    \begin{small}
    \boldsymbol{\Theta}(\boldsymbol{X}, \boldsymbol{U}) = 
    \begingroup 
    \setlength\arraycolsep{4pt}
    \begin{bmatrix}
        \; \; \vertbar \; \; & \vertbar    \; \; & \vertbar         \; \; &        \; \; & \vertbar \; \\
        \; \; 1        \; \; & \boldsymbol{X}  \; \; & \boldsymbol{X}^{P_2} \; \; & \cdots \; \; & \sin(\boldsymbol{U}) & \; \; \cdots \; \;  \\ 
        \; \; \vertbar \; \; & \vertbar    \; \; & \vertbar         \; \; &        \; \; & \vertbar \;
    \end{bmatrix}, 
    \endgroup 
    \label{eq:exp_Thetax}
    \end{small}
\end{equation}
where $\boldsymbol{X}^{P_2}$ 
%
denotes the quadratic
nonlinearities in the state variable $\boldsymbol{X} $. 
While we have chosen function library $\boldsymbol{\Theta}$ for the dynamical systems 
in our experiments, 
in practice, the $\boldsymbol{\Theta}$ could be expanded to include a wider range of nonlinear basis functions tailored to the system in question. 

\begin{figure*}[!t]
    \begin{center}
    \includegraphics[width=1\textwidth]{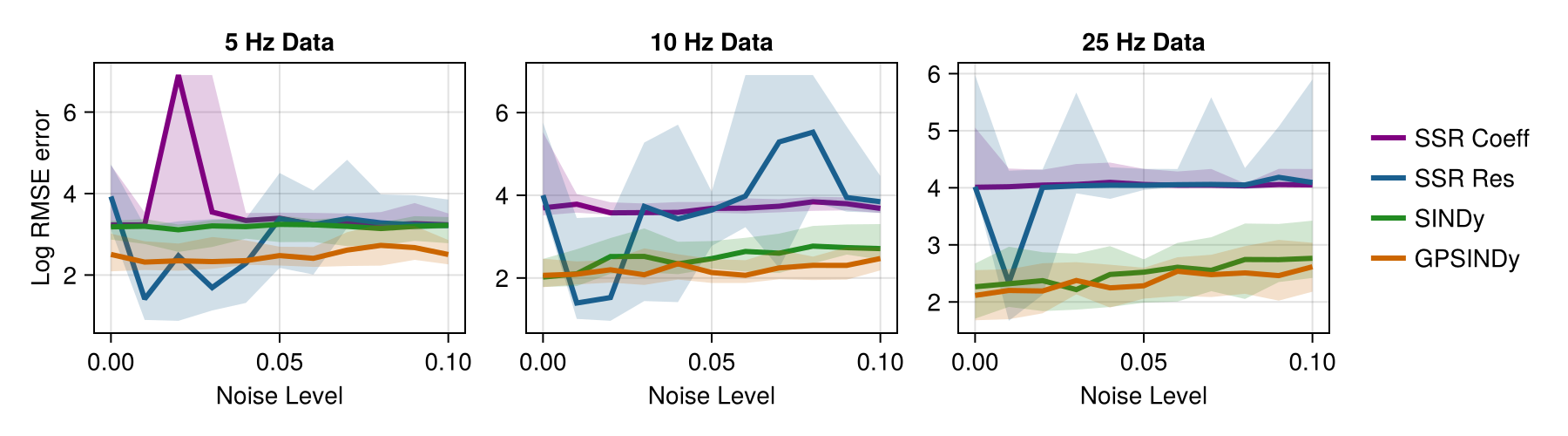}
    \end{center}
    \vspace{-2em}
    \caption{\small{\textbf{GPSINDy outperforms baselines on NVIDIA JetRacer trajectories under noisy measurements.} Each subplot represents the NVIDIA JetRacer dataset at different frequencies (Hz). 
    The horizontal axis represents the noise level standard deviations and the vertical axis the log root mean-squared error (RMSE) between the predicted trajectory (from learned system dynamics) and ground-truth states. 
    The ribbons indicate the upper and lower quartiles around the median $\log$ RMSE; lower overall error is better. 
    Over 
    45 rollouts, although SSR Residual (blue) sometimes beats GPSINDy (orange), GPSINDy achieves the lowest RMSE for the most frequencies and noise levels. 
    }} 
    \label{fig:combined_metrics}
\vspace{-2em}
\end{figure*}


\subsection{Lotka-Volterra Model (Predator/Prey)} 

We first consider the problem of identifying the equations of motion for the Lotka-Volterra model \cite{jorgensen_prey_predator}, which models the size of predator and prey populations over time 
\begin{align}
    \dot{x}_1 = a x_1 - b x_1 x_2,  \hspace{2em} \dot{x}_2 = -c x_2 + d x_1 x_2. 
    \label{eq:predator-prey}
\end{align}
The variable $x_1$ represents the prey population, $x_2$ the predator population, $a = 1.1$ the prey growth rate, $b = 0.4$ the effect of predation upon the prey population, $c = 1.0$ the predator's death rate, and $d = 0.4$ the growth of predators based on the prey population. 
For this experiment, we 
standardize the data, i.e. normalize it to have zero mean and unit variance, 
to improve parameter estimation of covariance functions and mitigate numerical issues associated with inverting ill-conditioned covariance matrices 
\cite{lingmont_thesis}. 

We first compare the learned coefficients  $\boldsymbol{\Xi}$ between SINDy and our proposed approach in Table \ref{tab:gpsindy_coeffs_compare}. 
We observe that the estimates for the parameters $a,b,c$, and $d$ obtained by GPSINDy are a closer approximation of the true underlying dynamics and that the coefficients matrix $ \boldsymbol{\Xi} $ learned by GPSINDy is also more sparse than the one learned by SINDy. 
This is because our approach uses GPs to estimate $ \dot{\boldsymbol{X}} $, which is a smoother approximation of the true derivatives of $ \boldsymbol{X}$ than the finite differenced derivatives.

We also quantitatively compare the results of SINDy, NNSINDy, and GPSINDy on data corrupted by different levels of noise as shown in Figure \ref{fig:predator_prey_unicycle_err_coeff}. The results show GPSINDy performs best across all noise magnitudes, highlighting its robustness in dealing with noisy data.
NNSINDy, constrained by limited data, fails to effectively learn model coefficients. 
SINDy is effective at low noise levels, but the error between the SINDy-learned dynamics and the ground truth dynamics increases with noise level.  
%


\subsection{Unicycle Dynamics (Simulation)} 

We now consider the unicycle system
\begin{equation}
\begin{aligned}
    \dot{x}_1 & = x_3 \cos(x_4), &
    \dot{x}_2 & = x_3 \sin(x_4), \\
    \dot{x}_3 & = u_1, &
    \dot{x}_4 & = u_2.
\end{aligned}
\end{equation}
We set the control inputs as $u_1(t) = \sin(t)$ and $u_2(t) = \frac{1}{2}\cos(t)$, 
deterministic functions that perturb the dynamics 
from an initial condition $\boldsymbol{x}_{0} = [0, 0, 0.5, 0.5]^{\top}$. 
In practice, any function can be chosen which perturbs the dynamics. 
Unlike the prior experiment, we opt to not standardize the states $ \boldsymbol{x}(t) $ since the data spread was already apt for Gaussian process regression. 
We note that we use polynomial terms up to $1^{\rm st}$ order in $ \boldsymbol{\Theta} ( \boldsymbol{X}_{GP}, \boldsymbol{U} ) $ as each method failed to identify the true coefficients when $3^{\rm rd}$ order terms were included. 

We compare SINDy, NNSINDy, and GPSINDy results on data with varying noise levels in Figure \ref{fig:predator_prey_unicycle_err_coeff}. 
While NNSINDy and GPSINDy perform similarly for both the predator-prey and unicycle systems, GPSINDy consistently outperforms other methods while SINDy notably degrades at higher noise levels. 
Although all methods show significant errors in learned coefficients compared to true system dynamics (suggesting model mismatch), 
%
our results show that GPSINDy learns more accurate coefficients even with more complex dynamics like the unicycle system and noisy measurements.

Additionally, this experiment demonstrates the impact of constructing the function library $\boldsymbol{\Theta}$ based on system knowledge. 
The true system dynamics only contain polynomial terms up to the $1^{\rm st}$ order, but when the library included terms up to the $ 3^{\rm rd  } $ order, each method overfit the model to the noise by assigning larger weights to the higher order terms. 
We also observed this effect in the predator-prey experiment but with greater impact on the unicycle system,
possibly due to approximating trigonometric functions in the true dynamics with Taylor series terms. 
Reducing the order of the library to $ 1^{ \rm st } $ order improved model fit for all methods. 

\subsection{JetRacer Hardware Demonstration} 

We also test our method on data collected from a NVIDIA JetRacer, a $1/10 
$ scale high speed car. 
We drove the car in a figure-8 made up of two circles, $3 \si{\meter}$ in diameter with nominal lap time of $5.5 \si{\second}$ and nominal velocity of $3.4 \si{\meter\per\second}$. 
VICON sensors captured $22.85 \si{\second}$ of the system's motion at discrete timesteps of $0.02 \si{\second}$ (50 Hz) with the control inputs $ \boldsymbol{U} $ saved at the same sampling rate for 45 total runs. 
We define the state $\boldsymbol{X}_i$ at time $t_i$ to be the measured $x_1$ and $x_2$ position in $\si{\meter}$, forward velocity magnitude $v$ of the car in $\si{\meter\per\second}$, and heading angle $ \phi $ (with respect to a global frame) in $\si{\radian\per\second}$. 
%
We stack each state measurement 
\eqref{eq:X_data} to gather $ \boldsymbol{X} $, after which we approximate $ \dot{\boldsymbol{X}} $ using central finite differencing. 

%
To evaluate performance for varying frequencies and noise levels, we downsample the 50 Hz state and control time histories to 25, 10, and 5 Hz and add noise $\boldsymbol{\epsilon} \sim \mathcal{N}\left(0, \sigma^2 \right)$ to the state measurements, with $\sigma$ varying from 0 to 0.1. 
We generate smoothed points $ \boldsymbol{X}_{GP} $ and $ \dot{\boldsymbol{X}}_{GP} $ at the same 
time points for each frequency. 
Finally, we compute $ \boldsymbol{\Theta}( \boldsymbol{X}_{GP}, \boldsymbol{U} ) $ 
%
and solve 
the 
$L_1$-regularized 
problem in \eqref{eq:admm} to obtain the GPSINDy dynamics model. 

\vspace{-0.5em}
\begin{table}[h!]
\centering 
\begin{tabular}{@{}lp{3.8mm}p{3.8mm}p{3.8mm}p{3.8mm}p{3.8mm}p{3.8mm}p{3.8mm}p{3.8mm}p{3.8mm}@{}}
\toprule
\textbf{}        & \multicolumn{3}{c}{\textbf{5 Hz }}            & \multicolumn{3}{c}{\textbf{10 Hz }}          & \multicolumn{3}{c}{\textbf{25 Hz }}           \\ 
\cmidrule(r){2-4} \cmidrule(lr){5-7} \cmidrule(l){8-10}  
\textbf{Method}  & \textbf{$\frac{1}{4}$Q}  & \textbf{M}    & \textbf{$\frac{3}{4}$Q}   & \textbf{$\frac{1}{4}$Q}  & \textbf{M}    & \textbf{$\frac{3}{4}$Q}   & \textbf{$\frac{1}{4}$Q}  & \textbf{M}    & \textbf{$\frac{3}{4}$Q}   \\ 
\midrule 
SSR Coeff        & 24.0         & 26.3          & 40.2          & 35.3         & 40.0         & 49.5          & 54.9         & 57.2          & 76.1          \\
SSR Res          & 6.1          & 24.6          & 54.9          & 7.5          & 46.0         & 284         & 51.7         & 57.1          & 141         \\ 
SINDy            & 15.6         & 24.3          & 29.4          & 8.4          & 13.0         & 20.8          & 7.2          & 12.1          & 20.8          \\
\textbf{GPSINDy} & \textbf{9.1} & \textbf{12.0} & \textbf{18.4} & \textbf{6.7} & \textbf{9.0} & \textbf{13.1} & \textbf{7.3} & \textbf{10.7} & \textbf{16.8} \\
\bottomrule
\end{tabular}
\caption{\small{\textbf{GPSINDy Achieves Lowest Overall Median RMSE Among Baselines 
on JetRacer Predicts  
Under Noisy Measurements.} 
This table shows the median (M), lower quantile ($\frac{1}{4}$Q), and upper quantile ($\frac{3}{4}$Q) RMSE across all noise levels for each 5, 10, and 25 Hz dataset. Lower is better. This table outlines our \textit{key contribution} that the GPSINDy method is more robust with noisy data, given its consistently low median and quantile RMSE metrics. 
}} 
\label{tab:mean_rmse_baselines}
\vspace{-1em}
\end{table}

\begin{figure}[!t]
    \centering
    \includegraphics[width = 0.45\textwidth,]{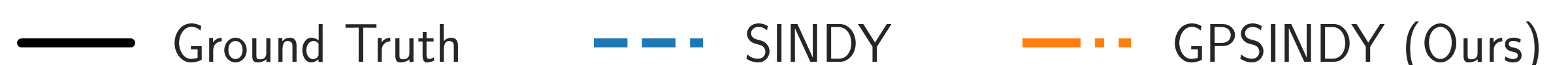}
    \includegraphics[width = 0.45\textwidth]{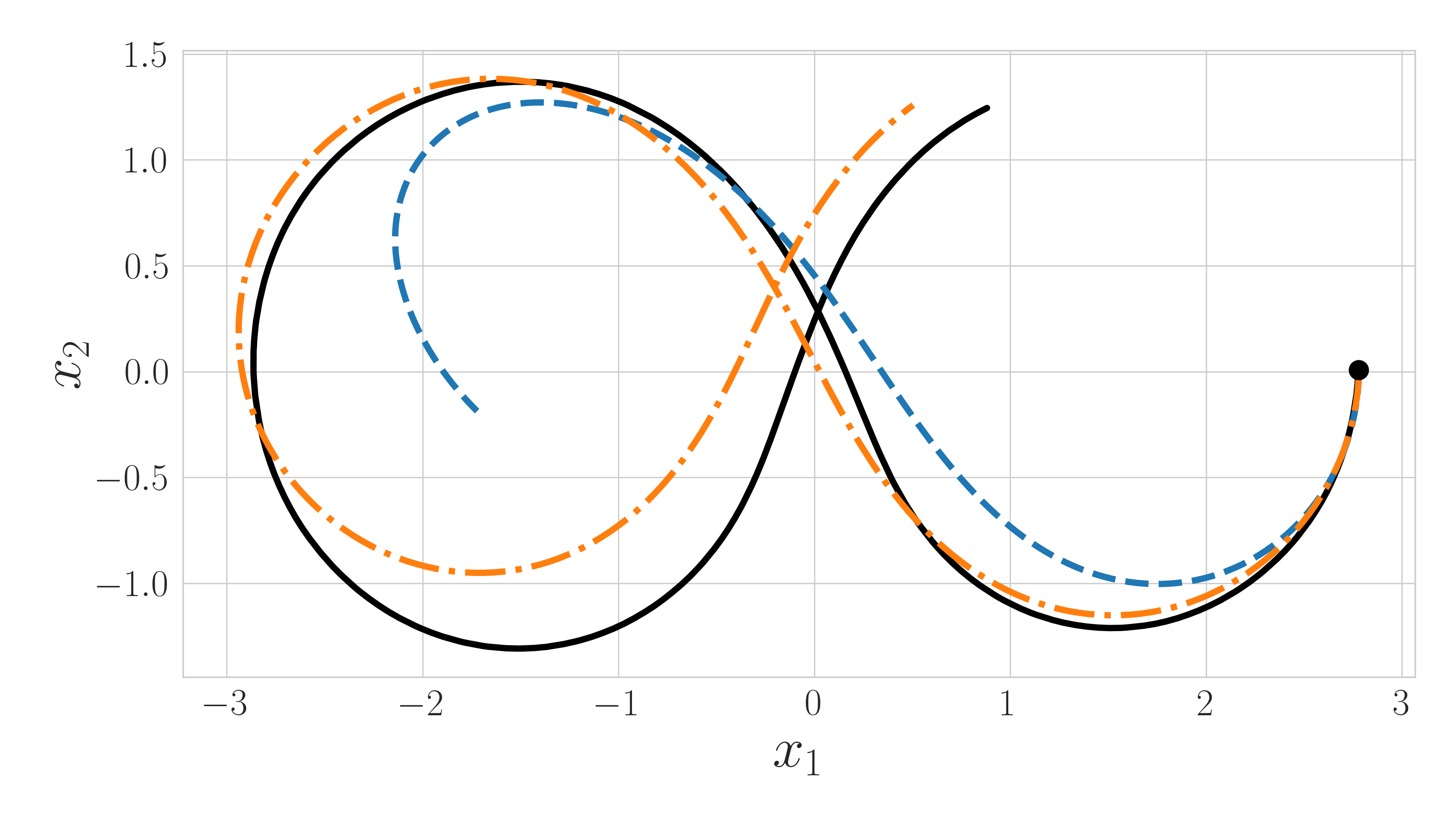}
    \vspace{-1em}
    \caption{
    \small{
    \textbf{GPSINDy Trajectories Align Closely with Ground Truth for the Real JetRacer System.} 
    Contrast the Cartesian trajectories predicted from SINDy (blue) and GPSINDy (orange) with the ground truth (black) based on one rollout out of the total collected JetRacer data. 
    For this trajectory, 
    the RMSE error norm between the $x_1$ and $x_2$ coordinates for SINDy on the testing data is $1.4\si{\meter\squared}$, while for GPSINDy it is reduced to $0.23\si{\meter\squared}$}. 
    } 
    \label{fig:single_jetracer}
    \vspace{-2em}
\end{figure}

Our baselines include 
Stepwise Sparse Regressor (SSR) Coefficient and Residual \cite{ssr_boninsegna2018sparse}, 
methods designed to mitigate noise in data like GPSINDy and serve as suitable benchmarks. 
SSR Coefficient (Coeff) truncates the smallest coefficient at each iteration while SSR Residual (Res) computes multiple models with varying sparsity, choosing the model with the lowest residual error. 
To achieve the best model fit, we tune $\lambda$ for each baseline via cross-validation, starting at $\lambda = 10^{-6}$ and increasing $\lambda$ by a factor of 10 until it reaches $1$. Then, we add 10 to $\lambda$ until all of the learned coefficients regress to 0. 
We propagate the dynamics for each $\lambda$ and, at the end, select the $\lambda$ for each $\dot{\boldsymbol{X}}$ that best fits the data. 
Figure \ref{fig:combined_metrics} displays the results for all 45 rollouts for GPSINDy and the other baselines. 
SSR Res shows the lowest error at low noise levels for 5 and 10 Hz, but GPSINDy achieves the lowest error overall across all frequencies as reported in Table \ref{tab:mean_rmse_baselines}. 
SSR specializes in identifying stochastic dynamical systems, which could explain its superior performance on some noisy data.  
However, the poor data quality still frequently leads to model mismatch, resulting in the large bands or quantile ribbons of the SSR Res error in Figure \ref{fig:combined_metrics} and error metrics in  
Table \ref{tab:mean_rmse_baselines}. 
GPSINDy predicts trajectories from data across all frequencies and noise levels with the lowest error bands among baselines. In particular, \textbf{ GPSINDy sees the largest gains when the data is sparse and noise-corrupted. } 
The overall median error for GPSINDy in Table \ref{tab:mean_rmse_baselines} is more than 50\% lower than other baselines for the 5 Hz data. 
%
%
Figure \ref{fig:single_jetracer} visually demonstrates the effectiveness of GPSINDy on one rollout of the original JetRacer dataset: The SINDy-predicted Cartesian trajectory initially aligns with the ground truth but increasingly deviates, while the GPSINDy model follows the ground truth trajectory. 

\section{Conclusions \& Future Work}
\label{sec:conclusions}

We have devised a method to 
learn models using limited data with high noise and sparsity in symbolic regression algorithms such as SINDy. 
We smooth sparse, noisy measurements using GP regression and then solve the LASSO problem with ADMM to learn more accurate dynamics over SINDy and other baselines. 
We demonstrate our approach on a Lotka-Volterra system 
and on noisy data taken from NVIDIA JetRacer hardware experiments. 
We show superior performance over baselines 
in predicting future trajectories from discovered dynamics using sparse and noisy data. 


%

\textbf{Future work: }
Future work could utilize the Bayesian framework of GPSINDy in an online model identification method, where each measurement updates the model in a sequential fashion, or explore using sparse Gaussian processes. 
%
%
Additionally, we should benchmark our method against directly taking the derivative of Gaussian Processes, provide a more thorough comparison between existing approaches, and conduct more testing on different dynamic systems. 


\newpage

\addtolength{\textheight}{-12cm}   


\bibliographystyle{ieeetr}
\bibliography{refs}

\end{document}